\documentclass[10pt,twocolumn,letterpaper]{article}

\usepackage{iccv}
\usepackage{times}
\usepackage{epsfig}
\usepackage{graphicx}
\usepackage{amsmath}
\usepackage{amssymb}

\usepackage{algorithmic}
\usepackage{algorithm}

\usepackage{mathrsfs}
\usepackage{color}
\usepackage{comment}

\usepackage{multirow}   
\usepackage{booktabs}   
\usepackage{tabularx}
\usepackage{ctable}
\usepackage{caption}
\usepackage{subfigure}

\newcommand{\mr}{\mathrm}
\newcommand{\mc}{\mathcal}

\newcommand{\RNum}[1]{\uppercase\expandafter{\romannumeral#1\relax}}

\usepackage[pagebackref=true,breaklinks=true,letterpaper=true,colorlinks,bookmarks=false]{hyperref} 

\iccvfinalcopy 

\def\iccvPaperID{****} 

\ificcvfinal\fi

\begin{document}

\title{Semantic-Transferable Weakly-Supervised Endoscopic Lesions Segmentation}
\author{
	Jiahua~Dong\textsuperscript{1,2,3},
	Yang Cong\textsuperscript{1,2}\thanks{The corresponding author is Prof. Yang Cong.}~,
	Gan Sun\textsuperscript{1,2,3}\thanks{The author contributed equally to this work.}~,
	Dongdong Hou\textsuperscript{1,2,3} \\	
	\textsuperscript{1}State Key Laboratory of Robotics, Shenyang Institute of Automation, \\ Chinese Academy of Sciences, Shenyang, 110016, China. \thanks{This work is supported by NSFC (61821005, 61722311, U1613214, 61533015), and LiaoNing Revitalization Talents Program (XLYC1807053). } \\	
	\textsuperscript{2}Institutes for Robotics and Intelligent Manufacturing, \\ Chinese Academy of Sciences, Shenyang, 110016, China. \\
	\textsuperscript{3}University of Chinese Academy of Sciences, Beijing, 100049, China. \\
	{\tt\small \{dongjiahua,~houdongdong\}@sia.cn~~\{congyang81,~sungan1412\}@gmail.com}
}

\maketitle
\ificcvfinal\thispagestyle{empty}\fi

\begin{abstract}													
Weakly-supervised learning under image-level labels supervision has been widely applied to semantic segmentation of medical lesions regions. However, 1) most existing models rely on effective constraints to explore the internal representation of lesions, which only produces inaccurate and coarse lesions regions; 2) they ignore the strong probabilistic dependencies between target lesions dataset (e.g., enteroscopy images) and well-to-annotated source diseases dataset (e.g., gastroscope images). To better utilize these dependencies, we present a new semantic lesions representation transfer model for weakly-supervised endoscopic lesions segmentation, which can exploit useful knowledge from relevant fully-labeled diseases segmentation task to enhance the performance of target weakly-labeled lesions segmentation task. More specifically, a pseudo label generator is proposed to leverage seed information to generate highly-confident pseudo pixel labels by incorporating class balance and super-pixel spatial prior. It can iteratively include more hard-to-transfer samples from weakly-labeled target dataset into training set. Afterwards, dynamically-searched feature centroids for same class among different datasets are aligned by accumulating previously-learned features. Meanwhile, adversarial learning is also employed in this paper, to narrow the gap between the lesions among different datasets in output space. Finally, we build a new medical endoscopic dataset with 3659 images collected from more than 1100 volunteers. Extensive experiments on our collected dataset and several benchmark datasets validate the effectiveness of our model.
\end{abstract}

\vspace{-1pt}
\section{Introduction}
Weakly-supervised learning \cite{exp:CDWS, Weak_super_1} focuses on learning a pixel-level lesion segmentation model for medical images with only weakly-labeled (image-level) annotations. Due to the slight requirements for large-scale, high-quality fully-labeled (pixel-level) annotations, it has been widely-explored in a number of medical diagnosis tasks, e.g., automated glaucoma detection \cite{zhao2019weakly}, thoracic disease localization \cite{yan2018weakly}, histopathology segmentation \cite{exp:CDWS}, etc.
\begin{figure}[t]    
	\small
	\centering
	\includegraphics[trim = 0mm 39mm 0mm 40mm, clip, width =237pt, height =110pt]{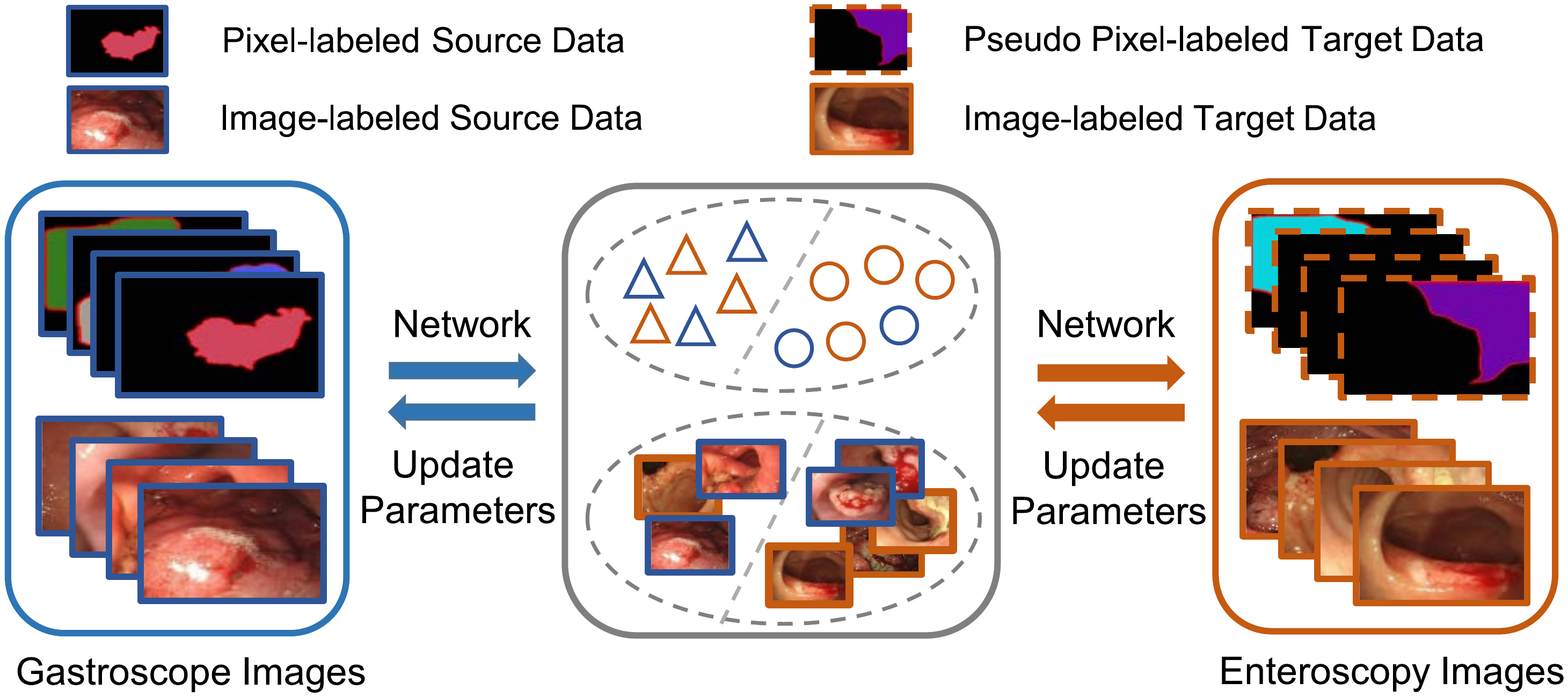} 
	\vspace{-16pt}  
	\caption{Demonstration of our semantic lesion representation transfer model, where the left and right images are from gastroscope and enteroscopy datasets, respectively. Our model learns the semantic transferable knowledge from source data to target data via pseudo pixel-label and dynamically-searched feature centroids (i.e., different shapes) of each class.}
	\label{figure:demonstration} 	
	\vspace{-15.0pt}
\end{figure}

However, weakly-supervised learning is a huge challenge for semantic lesions segmentation since 1) effective constraints or domain expertise are needed to learn the internal representation related to image-level annotations, which can produce inaccurate and coarse lesion regions; 2) it ignores the strong probabilistic dependencies between target lesions segmentation task and well-to-annotated source diseases, where such dependencies are treated as semantic knowledge. For example, diseases detected by both gastroscope and enteroscopy tend to share similar appearances, and further have similar prior distributions. Based on such dependencies, in this paper, we explore how to transfer semantic knowledge from closely-related fully-annotated source dataset (e.g., gastroscope images) to weakly-labeled target dataset (e.g., enteroscopy images).



To take advantage of the semantic transferable knowledge, we propose a new weakly-supervised semantic lesions representation transfer model as shown in Figure~\ref{figure:demonstration}, and its goal is to learn the transferable semantic knowledge from fully-labeled source diseases dataset to improve the segmentation performance on target weakly-labeled lesions segmentation task. The core idea of our model is a pseudo pixel-label generator, which can leverage seed information by incorporating class balance with super-pixel prior \cite{SLIC} to further prevent the  dominance of well-to-transfer categories. The hard-to-transfer samples can be incrementally introduced from the target dataset into training set. Afterwards, to mitigate the mapping features gaps of same class among source and target datasets, we endeavor to learn transferable knowledge by aligning the dynamically-searched feature centroids, which are gradually reckoned with previously-learned features and highly-confident pseudo labels. Meanwhile, adversarial learning is utilized in the output space to drive the segmentation outputs of the source and target datasets to share closer global distribution. Finally, we conduct the experiments on our built medical endoscopic dataset and several benchmark datasets to justify the superiority of our model. The experimental results can strongly support the effectiveness of our proposed model.


The contributions of our work are as follows:	
\vspace{-4pt} 		
\begin{itemize}
	\setlength{\itemsep}{1pt}		
	\setlength{\parsep}{0pt}	
	\setlength{\parskip}{0pt}											
	\item We develop a new semantic lesion representation transfer model for weakly-supervised lesions segmentation. To our best knowledge, this is an earlier exploration about semantic transfer for endoscopic lesions segmentation in the medical image analysis field.
	
	\item A pseudo pixel label generator is proposed to progressively mine more highly-confident pseudo labels, which can not only include more hard-to-transfer samples from the target dataset into training set, but also achieve class balance with super-pixel priors.
	
	\item A new medical endoscopic dataset with 3659 images collected from more than 1100 volunteers is built. We demonstrate the effectiveness of our model against several state-of-the-arts on our endoscopic dataset and several benchmark datasets. 	
	
\end{itemize}

\section{Related Work}	
In this section, we discuss some representative related works about semantic lesion segmentation and semantic representation transfer.

\textbf{Semantic Lesion Segmentation:} Computer aided diagnosis (CAD) \cite{cad_1, cui2019simultaneous,7407336,CONG2015907} is developed to assist clinician to improve the efficiency and accuracy of medical lesions segmentation. Traditional methods rely on local image features handcrafted by domain experts \cite{traditional:seg1, traditional:seg4}. To further improve the segmentation quality, most advanced methods \cite{medical_segment_springer, medical_seg_fcn, medical_fcn_1} based on convolutional neural networks \cite{related:resnet, related:vgg, net:deeplabv3} are proposed, which can achieve state-of-the-arts performance but acquire lots of pixel-level annotations. Thus, weakly-supervised semantic lesions segmentation methods \cite{exp:CDWS, Weak_super_1} are proposed to save annotation efforts. However, there is currently still a large segmentation performance gap between models trained only with image tags and models trained with pixel annotations.

\textbf{Semantic Knowledge Transfer:} Learning the semantic transferable representation from source dataset to target dataset for classification task via generative adversarial network \cite{Goodfellow:2014:GAN:2969033.2969125} has been widely-explored \cite{domain:class-Long, domian:class-Tzeng-1, domian:class-Tzeng-2, Long:2016:UDA:3157096.3157112, domain:class-preserve-2}. As pointed out in \cite{exp:CL}, methods addressing classification transfer do not translate well to the semantic segmentation task, which is still a significant challenge. Recently, Bousmalis et al. \cite{domain:transfer} propose to learn transferable knowledge via transferring the source images to target dataset. \cite{exp:CL} utilizes a curriculum learning approach to mitigate the gap between source and target dataset. Several researches \cite{DBLP:journals/corr/HoffmanWYD16, exp:CCA, domain:class-preserve-2,  Dou:2018:UCD:3304415.3304514, exp:LtA} focus on employing adversarial learning to semantic segmentation transfer in the feature space. \cite{exp:CGAN} introduces an additional generator conditioned on the extra auxiliary information for target dataset. \cite{Zou_2018_ECCV} exploits a self-training strategy for semantic representation transfer. However, existing models cannot be directly applied to semantic lesion transfer, since 1) they cannot ensure the features in same class but in different datasets are mapped nearby due to non-valid labeled information for target samples; 2) the model tends to transfer some easier-to-learned classes instead of balancing all the classes.

Therefore, we focus on learning semantic transferable knowledge by highly-confident class-balanced pseudo labels and dynamically-searched feature centroids with previously-learned experience.

\setlength{\belowcaptionskip}{-15pt}
\begin{figure*}[t]      
	\small
	\centering
	\includegraphics[trim = 0mm 60mm 0mm 60mm, clip, width =500pt, height =152pt]{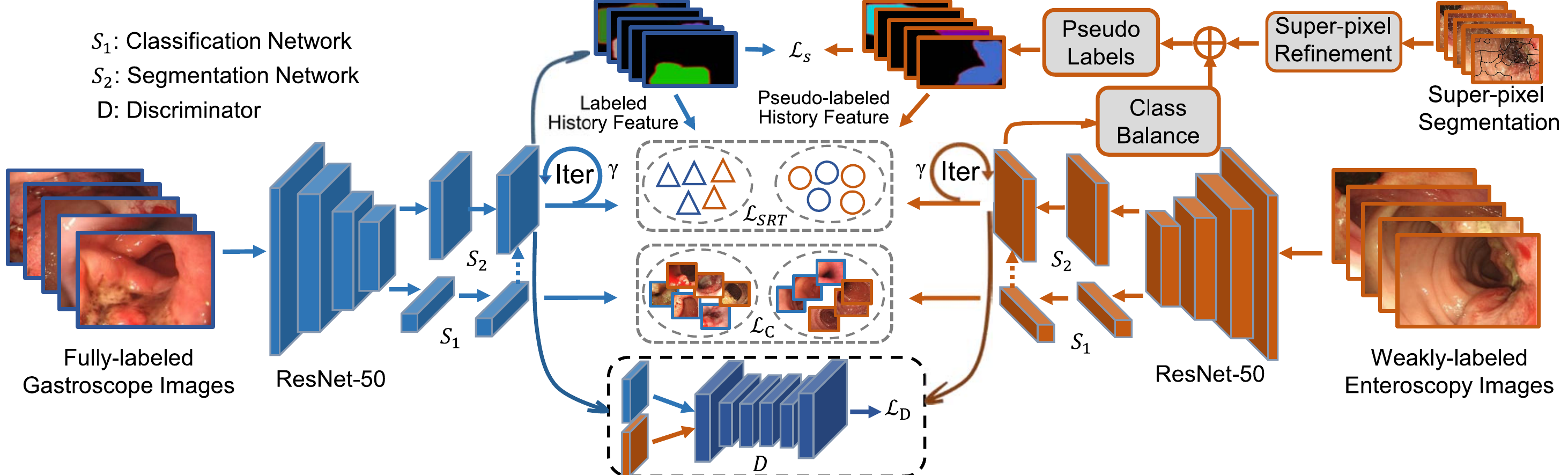}
	\vspace{-18.0pt}
	\caption{Framework of our proposed model, where six components of our model include \textit{ResNet-50} network for feature extraction, \textit{adversarial learning} for enforcing various lesion segmentation to share closer distribution, \textit{pseudo label generator} for weakly-labeled enteroscopy dataset, \textit{semantic representation transfer loss} $\mc{L}_{SRT}$ for aligning feature centroids among source and target datasets, and \textit{two subnets} denoted as $S_1$ and $S_2$ for classification $\mc{L}_C$ and segmentation $\mc{L}_S$, respectively.}
	\label{figure:overview_algorithm} 	
\end{figure*}		
\vspace{4pt}

\section{The Proposed Model}
In this section, we provide a brief overview about our semantic lesion representation transfer model. Then, the details about model formulation, training and testing procedures are elaborated.

\subsection{Overview of Our Proposed Model}				
The overview architecture of our model is shown in Figure~\ref{figure:overview_algorithm}. Two subnets marked as $S_1$ and $S_2$ are designed for classification and segmentation tasks, respectively, where the prediction of subnet $S_2$ is refined by classification probability via convolution operation, as shown in dashed arrows of Figure~\ref{figure:overview_algorithm}. Suppose that the source dataset (e.g., gastroscope images) and target dataset (e.g., enteroscopy images) are denoted as $X^s=\{(x_i^s,y_{i}^{sc},y_{i}^{ss})\}_{i=1}^{n_s}$ and $X^t=\{(x_j^t,y_{j}^{tc}\}_{j=1}^{n_t}$, respectively, where $y_{i}^{sc}$ and $y_{i}^{ss}$ are the corresponding image and pixel annotations of $x_i^s$, and $y_{j}^{tc}$ is the corresponding image annotation of $x_j^t$. We firstly forward image $x_i^s$ of source dataset $X^s$ to optimize the whole network excluding discriminator $D$. The segmentation output for image $x_j^t$ of target dataset $X^t$ is then predicted by subnet $S_2$. Since our goal is to encourage the segmentation outputs of source dataset $X^s$ and target dataset $X^t$ to share closer distribution, discriminator $D$ takes these two predictions as the input to distinguish whether the input is from $X^s$ or $X^t$.

Although we employ generative adversarial objective to narrow the gap of segmentation outputs between $X^s$ and $X^t$, it cannot ensure the features of same class in different datasets (i.e., $X^s$ and $X^t$) are mapped nearby. Inspired by this key observation, we endeavor to learn the semantic representation transfer by aligning the feature centroid for each class. However, we do not have pixel annotations as guidance to compute centroids for target dataset $X^t$. To address this issue, we propose a new method to generate pseudo pixel labels, which takes into account class balance and super-pixel segmentation priors. Based on the pseudo labels of target dataset, we utilize exponentially-weighted features based on previously-learned experience to compute semantic centroid for each class. Furthermore, the target image $x_j^t$ assigned with pseudo pixel labels $\hat{y}_{j}^{ts}$ is then forwarded into our model to fine-tuning the whole network.

\subsection{Model Formulation}

In order to learn transferable knowledge for target disease segmentation task, we formulate our proposed model as the following objective:	
\begin{equation}
\begin{aligned}
\mathcal{L} = &\mathcal{L}_C(X^s, X^t) + \mathcal{L}_S(X^s, X^t)+\eta\mathcal{L}_D(X^s, X^t) \\
&+ \mu\mathcal{L}_{SRT}(X^s, X^t),
\end{aligned}		
\label{eqution:loss_all}		
\end{equation}
where $\eta\geq 0$ and $\mu\geq 0$ are trade-off parameters and the definitions of each loss function are shown as follows:	

\textbf{Classification Loss $\mathcal{L}_C(X^s, X^t)$:} $\mathcal{L}_C(X^s, X^t)$ represents the classification loss of both target and source datasets (e.g., gastroscope and enteroscopy datasets). The subnet $S_1$ is utilized to discriminate whether the input image has lesion or not by the loss $\mathcal{L}_C(X^s, X^t)$:		
\begin{equation}
\begin{aligned}
\mathcal{L}_C(X^s, X^t) = &\mathbb{E}_{(x_i^s, y_{i}^{sc}) \in X^s}\big(J(S_1(x_i^s, \theta_{S_1}), y_{i}^{sc})\big)  \\
+&\mathbb{E}_{(x_j^t, y_{j}^{tc}) \in X^t}\big(J(S_1(x_j^t, \theta_{S_1}), y_{j}^{tc})\big),
\end{aligned}			
\label{equation:loss_classification} 								
\end{equation}
where $\theta_{S_1}$ denotes the parameters of the subnet $S_1$. $S_1(x_i^s, \theta_{S_1})$ and $S_1(x_j^t, \theta_{S_1})$ are the classification softmax outputs for source and target datasets, respectively, and $J(.,.)$ is the typical cross-entropy loss.		

\textbf{Segmentation Loss $\mathcal{L}_S(X^s, X^t)$:} For the subnet $S_2$ with softmax outputs, $\mathcal{L}_S(X^s, X^t)$ can be formulated as the segmentation loss for dataset $X^s$ with supervised pixel annotation $y_{i}^{ss}$, and dataset $X^t$ with assigned pseudo pixel label $\hat{y}_{j}^{ts}$. It can then be formulated as:		
\begin{equation}								
\begin{split}										
\mathcal{L}_S(X^s, & X^t)\! =\! \mathbb{E}_{(x_i^s, y_{i}^{ss}) \in X^s}\big(\!\!-\! \sum\limits_{a=1}^{|x_i^s|} (y_{ia}^{ss})^{\!\top} \log(S_2(x_i^s, \theta_{S_2})_a)\big)   \\
+ & \mathbb{E}_{x_{j}^t \in X^t} \big(\!\! -\! \sum\limits_{b=1}^{|x_{j}^t|} (\hat{y}_{jb}^{ts})^{\top} \log(S_2(x_j^t, \theta_{S_2})_b) \!+\! \lambda \left\| \hat{y}_{jb}^{ts}\right\|_1 \big),  \\				
s.t.,&\;\;  \hat{y}_{jb}^{ts} \in \big\{\{ \mathbf{e}_k| \mathbf{e}_k \in \mathbb{R}^K\} \cup \mathbf{0}\big\}, ~~\forall{b}=1,\ldots |x_{j}^t|,			
\end{split}	
\label{equation:loss_segmentation}				
\end{equation}		
where $\theta_{S_2}$ denotes the parameters of $S_2$, $S_2(x_i^s, \theta_{S_2})_a$ and $S_2(x_j^t, \theta_{S_2})_b$ are the segmentation softmax outputs of subnet $S_2$ at pixel $a$ ($a=1,2,...,\left| x_i^s \right|$) and $b$ ($b=1,2,...,\left| x_j^t \right|$), respectively. $y_{ia}^{ss}$ denotes one-hot encoding of ground truth label for the $a$-th pixel position in image $x_i^s$, and $\hat{y}_{jb}^{ts}$ is assigned pseudo label for the $b$-th pixel position in image $x_j^t$. $K$ and $\mathbf{e}_k$ are the number of classes and one-hot vector, respectively. Notice that assigning $\hat{y}_{jb}^{ts}$ as $\mathbf{0}$ can neglect this pseudo pixel label in training procedure. We thus expect the $\ell_1$-norm regularization on $\hat{y}_{j}^{ts}$ can serve as a negative sparse constraint to prevent the trivial solution from ignoring all pseudo pixel labels. $\lambda\geq 0$ is a global weight to control the amount of selected pseudo labels, and a larger $\lambda$ can promote the selection of more pseudo labels for model training.

Similar to self-paced learning \cite{kumar2010self}, Eq.~\eqref{equation:loss_segmentation} in our model can iteratively produce pseudo pixel labels corresponding to large confidence. However, the optimization of the second term in Eq.~\eqref{equation:loss_segmentation} can result in two issues: \textbf{(i)} our model will tend to be biased towards initially easily-learned classes and neglect other hard-to-transfer classes in the training procedure; \textbf{(ii)} the generated pseudo labels with highly-confident scores are spatially discrete. To address the issue \textbf{(i)}, the second term in Eq.~\eqref{equation:loss_segmentation} can be formulated as Eq.~\eqref{equation:optimization_of_pseudo_label} where class-wise confidence levels are normalized. 			
\begin{equation}
\begin{split}
&\min\limits_{\hat{y}_{jb}^{ts}} \mathbb{E}_{x_j^t}\big(\!-\! \sum\limits_{b=1}^{\left| x_j^t \right|} \sum\limits_{k=1}^{K} (\hat{y}_{jb}^{ts})_k \log(S_2(x_j^t, \theta_{S_2})_b) + \lambda_k \left\|\hat{y}_{jb}^{ts} \right\|_1\big),    \\  		
& s.t., \hat{y}_{jb}^{ts} = [(\hat{y}_{jb}^{ts})_1, ..., (\hat{y}_{jb}^{ts})_K] \in \big\{\{ \mathbf{e}_k| \mathbf{e}_k \in \mathbb{R}^K\} \cup \mathbf{0}\big\},
\end{split}
\label{equation:optimization_of_pseudo_label}
\end{equation}
where $\lambda_k$ ($k=1,2,...,K$) are class balance parameters that determine the proportion of generated pseudo labels for each class $k$. In order to avert dominance of large amount of pixel classes, we develop a new method for the determination of $\lambda_k$ as summarized in \textbf{Algorithm~1}: after obtaining maximum predicted probability $M_j$ of each pixel for all target images, we sort the probabilities of all pixels predicted as class $k$. $\lambda_k$ can be determined when $e^{- \lambda_k}$ equals to the probability ranked at $(1-p)\mr{length}(SM_k)$. The value of $p$ is starting from 25\% and empirically added by 5\% in each training epoch, and the maximum portion $p$ is set as 55\%. Furthermore, the optimal solution of Eq.~\eqref{equation:optimization_of_pseudo_label} is:
\begin{equation}
\begin{split}
(\hat{y}_{jb}^{ts})_k =
\left\{
\begin{aligned}		
&1, \quad \mr{if} ~~ k = \mathop{\arg\max}_{k} \frac{S_2(x_j^t, \theta_{S_2})_b}{e^{- \lambda_k}} ~~~\mr{and}   \\
& \qquad \quad  S_2(x_j^t, \theta_{S_2})_b > e^{- \lambda_k},  \\
&0, \quad\quad\quad\quad \mr{otherwise}.  \\
\end{aligned} 					
\right.											
\end{split}							
\label{equation:solution_of_pseudo_label}	
\end{equation}

To handle the issue \textbf{(ii)}, the pseudo labels that are produced by Eq.~\eqref{equation:solution_of_pseudo_label} can be refined with super-pixel spatial priors \cite{SLIC}, which ensures spatial continuity of generated pseudo labels. Moreover, \textbf{Algorithm~2} presents the details about how to apply super-pixel spatial refinement for the assignment of pseudo labels $\hat{y}_{j}^{ts}$: the super-pixel priors $S_j^t$ is applied for each target image $x_j^t$. When the $(h, w)$-th pixel which has same spatial priors among its 8-neighborhoods has no valid pseudo labels, its pixel label can be decided via voting the pseudo labels of its 8-neighborhoods.	

\textbf{Adversarial Loss $\mathcal{L}_D(X^s, X^t)$:} To drive lesion segmentation outputs between $X^s$ and $X^t$ to share similar distribution, we utilize generative adversarial objective $\mathcal{L}_D(X^s, X^t)$ in this paper. Discriminator $D$ in Figure~\ref{figure:overview_algorithm} takes these two segmentation softmax outputs of subnet $S_2$ as input to distinguish whether the input is from $X^s$ or $X^t$, and $S_2$ is trained to fool $D$. Formally, it can be defined as: 	
\begin{equation}
\begin{split}
\mathcal{L}_D(X^s, &X^t) = \mathbb{E}_{x_{j}^t \in X^t}\big(\log(D(S_2(x_j^t, \theta_{S_2}), \theta_D))\big)  \\	
& + \mathbb{E}_{x_{i}^s \in X^s}\big(\log (1 - D(S_2(x_i^s, \theta_{S_2}), \theta_D))\big),
\end{split}
\label{equation:loss_adv}	
\end{equation}
where $D(S_2(x_i^s, \theta_{S_2}), \theta_D)$ and $D(S_2(x_j^t, \theta_{S_2}), \theta_D)$ indicate the output of discriminator $D$ for image $x_i^s$ and $x_j^t$, respectively, and $\theta_D$ indicates the parameters of discriminator $D$.

\setlength{\textfloatsep}{7pt}
\renewcommand{\algorithmicrequire}{\textbf{Input:}}
\renewcommand{\algorithmicensure}{\textbf{Output:}}
\begin{algorithm}[t]			
	\caption{\small Determination of $\lambda_k$ in Eq.~\eqref{equation:optimization_of_pseudo_label}}
	\begin{algorithmic}[1]
		\REQUIRE Subnet $S_2$, the number of classes $K$, portion $p$ of selected pseudo labels, target image $x_j^t \in X^t$;
		\ENSURE $\lambda_k$    \\
		\FOR {$j=1,\ldots,\left|X^t\right|$}
		\STATE  Set $MP_k = \emptyset$;
		\STATE  $L_j = \mr{argmax}(S_2(x_j^t, \theta_{S_2}), \mr{axis}=3)$;
		\STATE  $M_j = \mr{max}(S_2(x_j^t, \theta_{S_2}),~\mr{axis}=3)$;
		\FOR {$k=1,\ldots,K$}
		\STATE  $M_{j}^k = M_j(L_j == k)$;
		\STATE  $MP_k = [MP_k, \mr{matrix\_to\_vector}(M_{j}^k)]$;
		\ENDFOR  \\
		\ENDFOR \\
		\FOR {$k=1,\ldots,K$}
		\STATE  $SM_k = \mr{sorting}(MP_k, \mr{ascending})$;
		\STATE  $T_k = (1 - p)\mr{length}(SM_k) $;
		\STATE  $\lambda_k = - \log(SM_k[T_k])$
		\ENDFOR \\
		return $\lambda_k$;
	\end{algorithmic}
\end{algorithm}							

\textbf{Semantic Transfer $\mathcal{L}_{SRT}(X^s, X^t)$:} To ensure that the features of same class in different datasets $X^s$ and $X^t$ are mapped nearby, $\mathcal{L}_{SRT}(X^s, X^t)$ is proposed for semantic representation transfer via feature centroid alignment, which can be defined as:
\begin{equation}	
\hspace{-5pt}\mathcal{L}_{SRT}(X^s, X^t) = \sum\limits_{k=1}^{K} \left\| C_k^s - C_k^t\right\|_2^2+\alpha\left\| C_k^s - C_k^t\right\|_1,	
\label{equation:loss_SRT}
\end{equation}
where $C_k^s$ and $C_k^t$ are the centroids of the class $k$ in datasets $X^s$ and $X^t$, respectively. $\alpha\geq 0$ is a trade-off parameter. Considering that the centroids of same class in different datasets have similar sparse property, we utilize the second term of Eq.~\eqref{equation:loss_SRT}. Specifically, motivated by exponential reward design in reinforcement learning \cite{RL-1, RL-2}, we propose a new method to search centroids for each class based on exponentially-weighted previously learned features, which resort to history learned experience. Furthermore, pseudo labels generated by \textbf{Algorithm 2} are used to guide semantic alignment for dataset $X^t$. The details of computing centroid for each class are shown in \textbf{Algorithm~3}.

Instead of aligning those newly obtained centroids in each iteration directly, we propose to align the centroids via resorting previously-leaned experience to overcome two practical limitations: 1) Categorical information in each batch is often insufficient, e.g., it is possible that some classes are missing in the current training batch since the samples are randomly selected; 2) If the batch size is small, even one false pseudo label will lead to the enormous deviation between the true centroid and pseudo-labeled centroid.

\renewcommand{\algorithmicrequire}{\textbf{Input:}}		
\renewcommand{\algorithmicensure}{\textbf{Output:}}		
\begin{algorithm}[t]	
	\caption{\small Determination of Ultimate Pseudo Pixel Labels}
	\begin{algorithmic}[1]	
		\REQUIRE Enteroscopy image $x_j^t \in X^t$, width $W$ and height $H$ of image $x_j^t$, the number of classes $K$;
		\ENSURE Pseudo labels $\hat{y}_{j}^{ts}$;   \\
		\STATE  Solve $\lambda_k$ via \textbf{Algorithm 1};
		\FOR {$j=1,\ldots,\left|X^t\right|~$}
		\STATE  Compute initial pseudo labels $\hat{y}_{j}^{ts}$ via Eq.~\eqref{equation:solution_of_pseudo_label};
		\STATE  Compute super-pixel segmentation priors $S_{j}^t$ of $x_j^t$;
		\FOR {$h=1,\ldots,H,~w=1,\ldots,W~$}
		\STATE  Set $C_{hw} = \emptyset$;
		\IF {$\hat{y}_{j}^{ts}$ has no pseudo labels at $(h,w)$-th pixel}	
		\FOR {$k=1,\ldots,K~$} 		
		\STATE  $C_{hw}^k\!\! =\!\! \sum\limits_{x= h-1}^{h+1}\!\sum\limits_{y=w-1}^{w+1} \!\!\!\mathbf{1}_{ \big( ((\hat{y}_{j}^{ts})_{xy} = k)\!\&((S_j^t)_{hw} = (S_j^t)_{xy})\big)}$			
		\STATE  $C_{hw} = [C_{hw}, ~C_{hw}^k]$;
		\ENDFOR
		\STATE  $N_k = \mr{argmax}(C_{hw}, ~\mr{axis}=0)$;
		\IF {$C_{hw}[N_k]~>~4~$}
		\STATE  $(\hat{y}_{j}^{ts})_{hw} = N_k$;
		\ENDIF
		\ENDIF
		\ENDFOR  \\ 	
		Return the ultimate pseudo labels $\hat{y}_{j}^{ts}$ for $x_j^t$;
		\ENDFOR  \\		
	\end{algorithmic}					
\end{algorithm}
\subsection{Details of Network Architecture}

\textbf{Baseline, Subnet $S_1$ and $S_2$:} 				
We utilize DeepLab-v3 \cite{net:deeplabv3} architecture based on ResNet-50 \cite{related:resnet} as the backbone network, which is pre-trained with ImageNet \cite{data:image-net}. For the ResNet-50 \cite{related:resnet}, we remove the last classification layer and modify the stride of the last two convolutional blocks from 2 to 1 for higher dimensional output. Moreover, three dilated convolutional filters with stride of \{1, 2, 4\} are utilized in the last convolutional block to enlarge receptive field. As shown in Figure~\ref{figure:overview_algorithm}, the output feature map generated by baseline ResNet-50 is passed into subnet $S_1$ for image classification. It is forwarded into subnet $S_2$ as well for pixel segmentation, which contains an Atrous Spatial Pyramid Pooling(ASPP) \cite{intro:deeplab} block and a pixel classifier layer.

\textbf{Discriminator (D):}
Inspired by \cite{DBLP:journals/corr/RadfordMC15}, for the discriminator $D$, we employ a fully convolutional networks for retaining global information compared with multi-layer perception. It consists of 5 convolutional layers with stride of 2 and kernel of 3. In more detail, the channels of 5 convolutional filters are \{16, 32, 64, 64, 1\}, respectively. Excluding the last convolution layer, the activation function of each filter is Leaky RELU with the parameter as 0.2.

\renewcommand{\algorithmicrequire}{\textbf{Input:}}
\renewcommand{\algorithmicensure}{\textbf{Output:}}
\begin{algorithm}[t]
	\caption{\small Optimizing Semantic Representation Transfer Loss}
	\begin{algorithmic}[1]
		\REQUIRE Max-iteration $N$, classes number $K$, the feature centroids $\{C_{k}^{s}\}_{k=1}^K$ and $\{C_k^t\}_{k=1}^K$ of each class $k$ for $X^s$ and $X^t$;
		\ENSURE $\mathcal{L}_{SRT}(X^s, X^t)$;  \\	
		\FOR {$n=1,\ldots,N$}
		\STATE  $\mathcal{L}_{SRT}(X^s, X^t) = 0$;
		\STATE  $\big((x_i^s, y_{i}^{ss}), (x_j^t)\big) = \mr{RandomlySampling}(X^s, X^t)$;
		\STATE  $\hat{y}_{j}^{ts} = \mr{PseudoLabeling}(x_j^t)$ via \textbf{Algorithm 2};
		\STATE  Extracting pixel feature maps $F_i^s$ and $F_j^t$ by subnet $S_2$ for $x_i^s\in X^s$ and $x_j^t\in X^t$	
		\FOR {$k=1,\ldots,K$}	
		\STATE  $C_{k}^{sn} = \frac{1}{\left|x_i^s \right|} \sum\limits_{a=1}^{\left|x_i^s \right|} (F_i^s)_{a} \textbf{1}_{(y_{i}^{ss})_{a}=k}$;
		\STATE  $C_{k}^{tn} = \frac{1}{\left|x_j^t \right|} \sum\limits_{b=1}^{\left|x_j^t \right|} (F_j^t)_{b} \textbf{1}_{(\hat{y}_{j}^{ts})_{b}=k}$;
		\STATE  $C_k^s\! =\! \sum_{x=1}^{n} C_{k}^{sx} \cdot \gamma ^{n-x}$; (Exponentially-weighted)
		\STATE  $C_k^t\! =\! \sum_{x=1}^{n} C_k^{tx} \cdot \gamma ^{n-x}$; (Exponentially-weighted)
		\ENDFOR   \\
		\STATE  Return $\mathcal{L}_{SRT}(X^s, X^t)$;
		\ENDFOR   \\
	\end{algorithmic}	
\end{algorithm}

\setlength{\belowcaptionskip}{-12pt}
\begin{table*}[t]
	\centering
	\setlength{\tabcolsep}{2.0mm}
	\begin{tabular}{|c|c|c|c|c|c|c|c|c|c|c|}
		\hline
		Metrics & Baseline \cite{net:deeplabv3} & CDWS \cite{exp:CDWS} & NMD \cite{exp:CCA} & Wild \cite{DBLP:journals/corr/HoffmanWYD16} & DFN \cite{exp:DFN} & LtA \cite{exp:LtA} & CGAN \cite{exp:CGAN} & Ours  \\
		\hline
		\hline
		$\mr{IoU}_n$($\%$) & 75.13 &25.11  & 81.10 & 81.58& 81.33 & 81.73& 80.32 & \textbf{84.76} \\	
		\hline
		$\mr{IoU}_d$($\%$) & 33.24  &15.51 &36.85& 38.59 &37.50&41.10 & 41.33 &\textbf{43.16}\\
		\hline
		mIoU($\%$) & 54.19  &20.31 &58.97 & 60.09 &59.41&61.42 & 60.82 & \textbf{63.96} \\
		\hline				
		
	\end{tabular}				
	\vspace{-5pt}
	\caption{Comparison performance between our proposed model and the state-of-the-arts on our medical dataset. Models with the best performance are bolded.}
	\label{table:exp_medical_dataset}
	\vspace{-2pt} 
\end{table*}

\subsection{Training and Testing}
\textbf{Training:} In each training step, for losses $\mathcal{L}_C(X^s, X^t)$ and $\mathcal{L}_S(X^s, X^t)$, we firstly forward the source image $x_i^s$ (e.g., gastroscope) with the image-level label $y_{i}^{sc}$ and the pixel-level annotation $y_{i}^{ss}$ to the network and generate the segmentation softmax output $S_2(x_i^s, \theta_{S_2})$. We then obtain the target softmax output $S_2(x_j^t, \theta_{S_2})$ for image $x_j^t$ (e.g., enteroscopy) only with the image-level label $y_{j}^{tc}$, and ultimate pseudo pixel labels $\hat{y}_{j}^{ts}$ are generated via \textbf{Algorithm~2}. In addition, these two segmentation outputs are passed into discriminator $D$ for optimizing $\mathcal{L}_D(X^s, X^t)$. For training the objective $\mathcal{L}_{SRT}(X^s, X^t)$, the centroids $C_k^s$ and $C_k^t$ for each class $k$ are computed via \textbf{Algorithm~3}, which resorts to previously learned features.

\textbf{Testing:} In testing phase, a target image $x_j^t$ (e.g., enteroscopy) is passed into feature extractor ResNet-50 followed by subnet $S_1$ and $S_2$ for classification and segmentation. The discriminator $D$ and other algorithmic designs would not be involved. As for implementation details, we use a single Titan XP GPU with 12 GB memory. The Adam optimizer is used to train whole networks with the batch size as 4. The initial learning rate is set as $1.0\times10^{-4}$ and it is exponential decay with the rate and step size as 0.7 and 950, respectively.

\section{Experiments}
In this section, we give detailed descriptions about our built dataset, and both source code and built dataset are available at \url{http://ai.sia.cn/lwfb/}. Although our model is mainly designed for medical image analysis, the experiments on other benchmark datasets are also conducted to validate its generalization performance.

\subsection{Dataset and Evaluation}
The datasets in our experiments include one our own medical dataset, and three benchmark datasets. 	

\textbf{Medical Endoscopic Dataset}: this dataset is built by ourself, which has total 3659 images that collected from more than 1100 volunteers with various lesions, including gastritis, polyp, cancer, bleeding and ulcer. Specifically, it contains 2969 gasteroscope images and 690 enteroscopy images. In the training phase, we treat the gasteroscope images as the source dataset, whose 2400 images have the image-level labels and 569 images have both image-level labels and pixel-level annotations; enteroscopy images are treated as target dataset, whose 300 images are with their image-level labels. For the test phase, the other 390 enteroscopy images are utilized to evaluate the performance.

\textbf{Cityscapes} \cite{data:city} is a real-world dataset about urban street scenes, which is collected in 50 cities. It consists of three disjoint subsets: training subset with 2993 images, validation subset with 503 images and test subset with 1531 images. There are total 34 distinct categories in the dataset. 	

\textbf{GTA} \cite{data:GTA} contains 24996 images w.r.t synthetic street scenes, which are collected from realistic computer game Grand Theft Auto V based on the city of Los Angeles. The segmentation annotations are compatible with the Cityscapes dataset \cite{data:city}.

\textbf{SYNTHIA} \cite{data:synthia} is a large synthetic dataset whose images are collected in virtual city without corresponding to any real city. For the experiments, we use its subset called SYNTHIA-RANDCITYSCAPES with 9400 images, including 12 automatically annotated object categories and some unnamed classes.

For the evaluation, we use intersection over union (IoU) as basic metric. Additionally, three derived metrics are also used, i.e., IoU of normal ($\mr{IoU}_n$), IoU of disease ($\mr{IoU}_d$) and mean IoU (mIoU). The larger of the corresponding metric is, the better of the corresponding model will be.

\subsection{Experiments on Medical Endoscopic Dataset}
In this subsection, we validate the superiority of our model by comparing it with several state-of-the-arts on our built medical dataset:
\vspace{-5pt}
\begin{itemize} 	
	\setlength{\itemsep}{0pt}
	\setlength{\parsep}{0pt}
	\setlength{\parskip}{0pt}
	\item{\textbf{Baseline}} (\textbf{BL}) model utilizes DeepLab-v3 \cite{net:deeplabv3} as backbone for segmentation without semantic transfer.
	
	\item{Constrained Deep Weak Supervision (\textbf{CDWS})} \cite{exp:CDWS} exploits multi-scale learning with weak supervision by applying area constraint for segmentation predictions.
	
	\item{No More Discrimination (\textbf{NMD})} \cite{exp:CCA} refines segmentation module by leveraging soft pseudo labels and static object priors with multiple class-wise adaptation.
	
	\item{FCNs in the Wild (\textbf{Wild})} \cite{DBLP:journals/corr/HoffmanWYD16} designs a adversarial loss with prior constraint on pixel-level output to optimize intermediate convolutional layers.
	
	\item{Discriminative Feature Network (\textbf{DFN})} \cite{exp:DFN} designs both Smooth Network and Border Network to learn discriminative semantic feature.		
	
	\item{Learning to Adapt (\textbf{LtA})} \cite{exp:LtA} exploits multi-level adaptation in the context of semantic segmentation.
	
	\item{Conditional GAN (\textbf{CGAN})} \cite{exp:CGAN} proposes to integrate conditional GAN into the segmentation network for feature space adaptation.	
\end{itemize}

\vspace{-5pt}
For a fair comparison, we use ResNet-50 \cite{related:resnet} as the backbone architecture and add an additional classification head to refine segmentation in this experiment. The evaluation results of our model against state-of-the-arts are presented in Table~\ref{table:exp_medical_dataset}. As shown in  Table~\ref{table:exp_medical_dataset}, we have the following observations: 1) Compared with the state-of-the-arts \cite{exp:LtA,exp:CGAN}, our proposed model outperforms them by a large margin around 2.54$\sim$3.14\%, which validates the effectiveness of our model, i.e., a pseudo label generator can mine more accurate and highly-confident pseudo labels. 2) As for mIoU, all models \cite{exp:CCA, DBLP:journals/corr/HoffmanWYD16, exp:DFN, exp:LtA, exp:CGAN} with semantic transfer outperform the models \cite{net:deeplabv3, exp:CDWS} without semantic transfer.


\begin{figure*}[t]      
	\small
	\centering
	\includegraphics[trim = 0mm 77mm 0mm 77mm, clip, width=450pt, height =87pt]{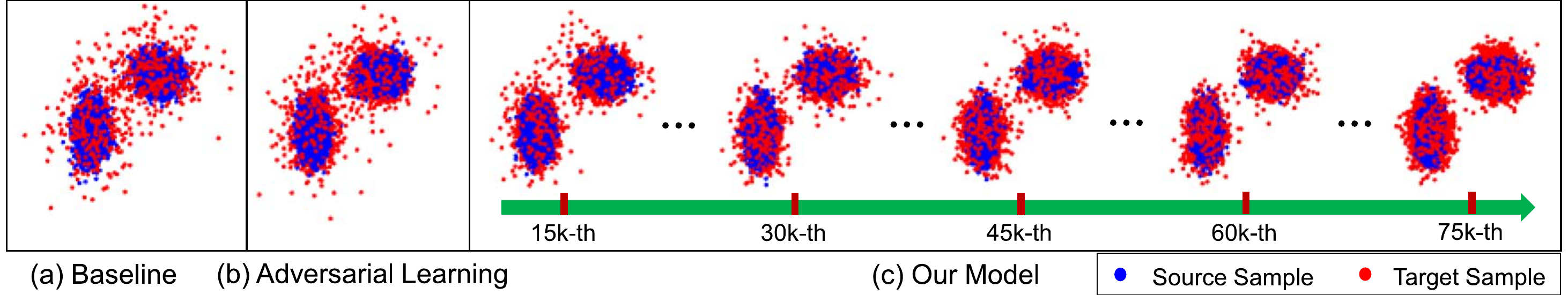}
	\hspace{0.4mm}
	\vspace{-8.0pt}
	\caption{Visualization of the learned representations using t-SNE \cite{t-SNE}, where blue and red points are source gasteroscope samples and target enteroscopy samples, respectively. Two separated clusters denote two categories, i.e., lesion and normal.} 	
	\label{figure:visual_learn_process} 	
\end{figure*}

\begin{table*}[ht]
	\vspace{6pt}
	\small
	\centering
	\setlength{\tabcolsep}{2.3mm}
  \scalebox{0.974}{
	\begin{tabular}{|c|c|c|c|c|c|c|c|c|c|}	
		\hline
		Metrics& BL & BL+AL & BL+AL+PL & BL+AL+SRT & BL+PL+SRT & Ours &Ours-woPL & Ours-woCB &  Ours-woSP  \\
		\hline
		\hline
		$\mr{IoU_n}$($\%$) &75.13 & 79.81 &83.08 &81.71 &84.38 &84.76 &81.71 &84.08 &84.22\\
		\hline		
		$\mr{IoU_d}$($\%$) & 33.24&39.27 &41.07 &41.27 &43.33 &43.16 &41.27 &40.51 &42.37  \\
		\hline		
		mIoU($\%$) & 54.19 & 59.54 &62.07 &61.49 &63.58 &63.96 & 61.69 &62.29 &63.30  \\		
		\hline
	\end{tabular}
     }		
	\vspace{-6pt}		
	\caption{Ablation study and different pseudo labels designs of our model on medical dataset with Baseline network DeepLab-v3 \cite{net:deeplabv3} (BL), Adversarial Learning (AL), Pseudo Labels (PL), Semantic Representation Transfer (SRT) and training without pseudo labels (Ours-woPL), class balance (Ours-woCB) or super-pixel spatial priors (Ours-woSP).}	
	\label{table:ablation_exp_medical}
	\vspace{-2pt}	
\end{table*}

\textbf{Ablation Study:} To validate the effectiveness of different components of our model, we also conduct experiment on our medical dataset with different components ablation, i.e., Baseline network DeepLab-v3 (BL), Adversarial Learning (AL), Pseudo Labels (PL) and Semantic Representation Transfer (SRT). As the results shown in Table~\ref{table:ablation_exp_medical}, we can observe that when one or more components are removed, the performance degrades, e.g., the performance decreases 0.38\%$\sim$4.42\% in terms of mIoU after removing the pseudo labels selection or semantic representation transfer. In addition, we also demonstrate the learned transferable representations in Figure~\ref{figure:visual_learn_process}. Notice that our model can well map the features of same class in different datasets nearby along the learning process when compared with Baseline (Figure~\ref{figure:visual_learn_process} (a)) and Adversarial Learning (Figure~\ref{figure:visual_learn_process} (b)), which validates that highly-confident pseudo pixel labels and previously-learned feature can further improve the performance for enteroscopy lesions segmentation.

\begin{figure}[t]
	\small
	\centering
	\includegraphics[trim = 0mm 45mm 0mm 45mm, clip, width =246.5pt, height =97pt]{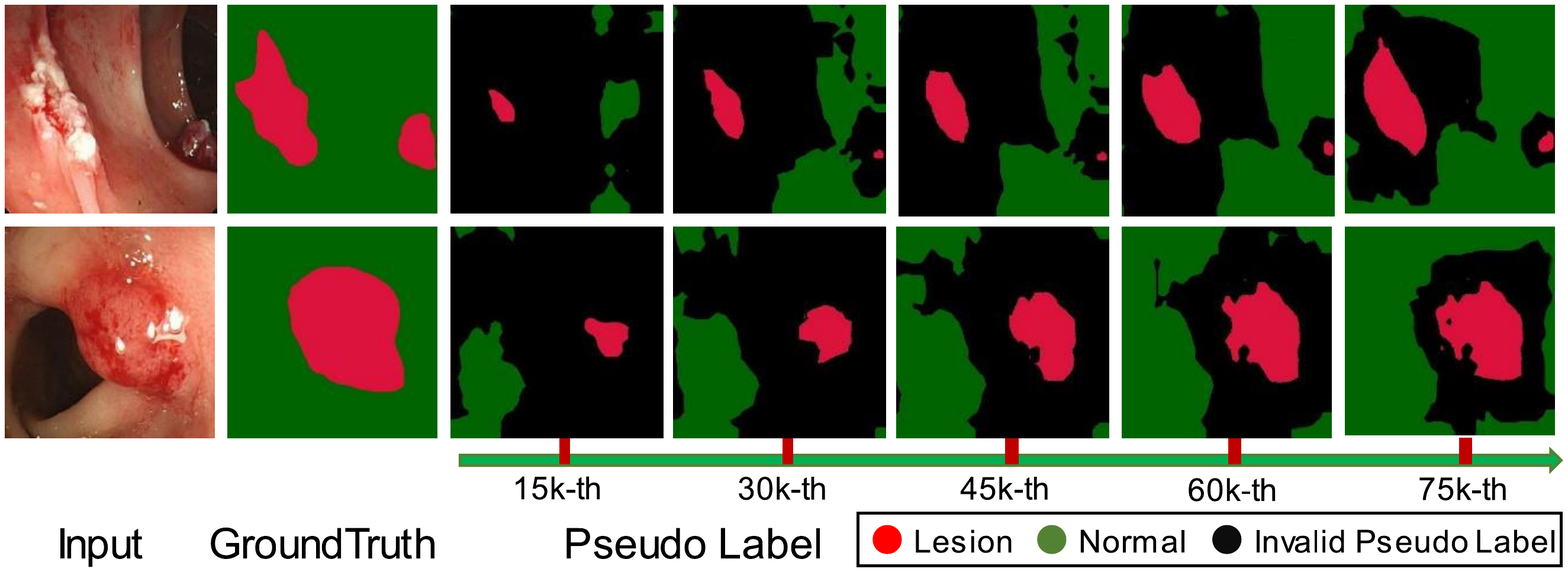}
	\vspace{-28.0pt}
	\caption{The illustration of intuitive propagation of pseudo labels, where input images are from enteroscopy dataset.}
	\label{fig:pseudo_label_propagation} 	
	\vspace{5pt}
\end{figure}

\textbf{Effect of Pseudo Labels Selection:} We intend to study how different designs for pseudo labels selection affect the performance of our model, i.e., training without pseudo labels (denoted as Ours-woPL), training without class balance (denoted as Ours-woCB) and training without super-pixel spatial priors (denoted as Ours-woSP). As the results shown in Table~\ref{table:ablation_exp_medical}, our model which is only with class balance can achieve $1.61\%$ improvement when comparing with Ours-woPL, while the training model with both class balance and super-pixel spatial priors can improve $2.27\%$. This observation indicates that the pseudo labels component is designed reasonably. In addition, as depicted in Figure~\ref{fig:pseudo_label_propagation}, the pseudo pixel label generator can iteratively generate more highly-confident pseudo pixel labels by incorporating class balance and super-pixel spatial prior.


\vspace{-2pt}

\begin{figure}[t]
	\vspace{-5pt}
	\begin{minipage}[t]{0.495\linewidth}
		\centering
		\includegraphics[trim = 38mm 75mm 90mm 150mm, clip, height=3.2cm,width=3.8cm]{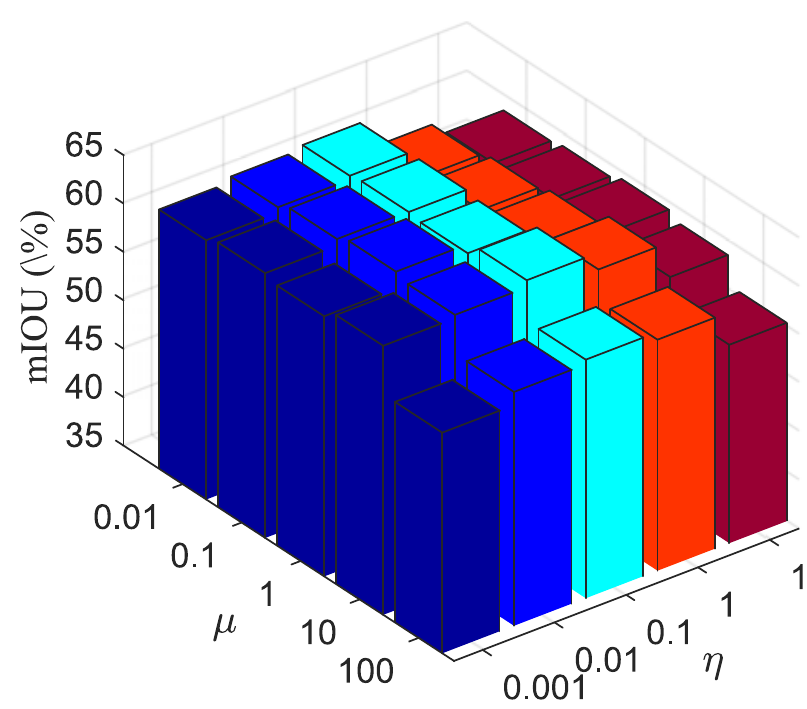}
		\vspace{-6pt}
		\caption*{(a) when $\gamma=0.7, \alpha=1$}
		\label{fig:paras_a}
	\end{minipage}
	\begin{minipage}[t]{0.495\linewidth}
		\centering
		\includegraphics[trim = 38mm 75mm 90mm 150mm, clip, height=3.2cm,width=3.8cm]{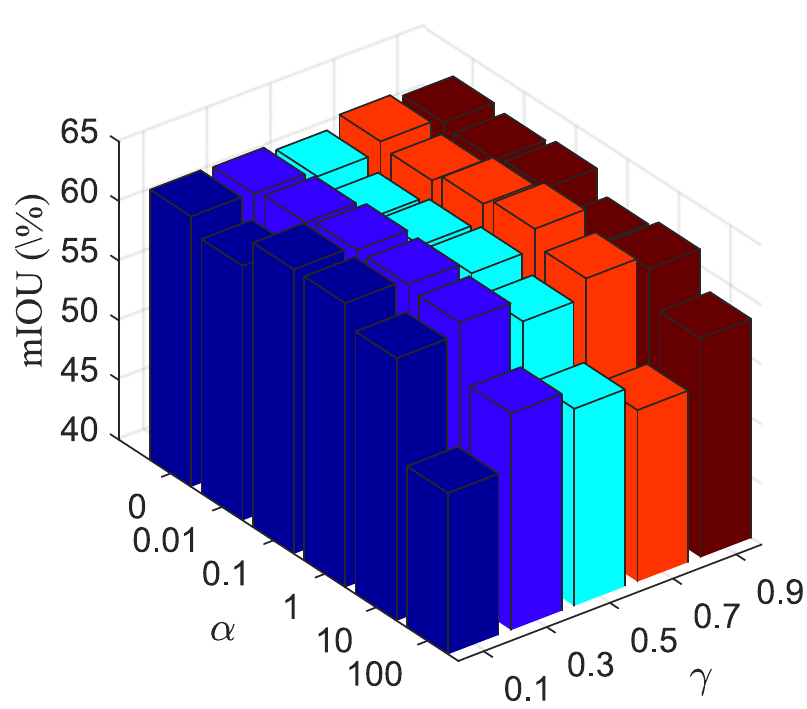}
		\vspace{-6pt}
		\caption*{(b) when $\mu=10, \eta=0.3$}
		\label{fig:paras_b}
	\end{minipage}
	\caption{The effect of parameters \{$\mu, \eta$\} (left) and \{$\alpha, \gamma$\} (right) on medical endoscopic dataset.}
	\label{fig:effect_paras}
	\vspace{4.5pt}
\end{figure}

\textbf{Effect of Hyper-Parameters:} This subsection investigates the effect of parameters $\{\mu, \eta\}$ and $\{\alpha, \gamma\}$. As the results illustrated in Figure~\ref{fig:effect_paras}, we can choose the optimal $\{\mu, \eta\}$ and $\{\alpha, \gamma\}$ by empirically conducting extensive parameter experiments. Notice that the performance of our model has great stability when tuning the value of different parameters. Moreover, it also validates the importance of incorporating previously-learned features and sparsity property of medical endoscopic dataset.

\begin{table*}[t]	
	\scriptsize
	\centering
	\setlength{\tabcolsep}{1.87mm}
	\begin{tabular}{|c|c|c|c|c|c|c|c|c|c|c|c|c|c|c|c|c|c|c|c|c|}
		\hline
		Method & road & sidewalk & building & wall & fence & pole &light &sign & veg & sky & person & rider & car & bus & mbike & bike & $\mr{mIoU}$ \\
		\hline			
		\hline			
		DF \cite{exp:DF} & 6.4 & 17.7 & 29.7 & 1.2 & 0.0 & 15.1& 0.0 & 7.2 &30.3 & 66.8 & 51.1 & 1.5 & 47.3 & 3.9 & 0.1 & 0.0 & 17.4 \\
		\hline
		Wild \cite{DBLP:journals/corr/HoffmanWYD16} & 11.5 & 19.6 & 30.8 & 4.4 & 0.0 & 20.3 & 0.1 & 11.7& 42.3 & 68.7 & 51.2 & 3.8 & 54.0 & 3.2 & 0.2 & 0.6 & 20.2  \\
		\hline
		CL \cite{exp:CL} & 65.2 & 26.1 & 74.9 & 0.1 & 0.5 & 10.7& 3.7 & 3.0 &76.1 & 70.6 & 47.1 & 8.2 & 43.2 & 20.7 & 0.7 & 13.1 & 29.0 \\ 	
		\hline
		NMD \cite{exp:CCA} & 62.7 & 25.6 & \textcolor{red}{\textbf{78.3}} & - & - & - & 1.2 & 5.4 & \textcolor{red}{\textbf{81.3}} & 81.0 & 37.4 & 6.4 & 63.5 &10.1 & 1.2 & 4.6 & - \\
		\hline
		LSD \cite{exp:LSD} &80.1 &29.1 &77.5 &2.8 &0.4 &26.8 &11.1 &18.0 &78.1 &76.7 &48.2 & 15.2&70.5 &17.4 &8.7 &16.7 &36.1 \\
		\hline
		LtA \cite{exp:LtA} & \textcolor{blue}{84.3} &  \textcolor{red}{\textbf{42.7}} & \textcolor{blue}{77.5}& -& -& -& 4.7 & 7.0& 77.9 & \textcolor{blue}{82.5} & 54.3 & 21.0 & 72.3 & 32.2 & 18.9 & 32.3 & - \\
		\hline	
		CGAN \cite{exp:CGAN} & \textcolor{red}{\textbf{85.0}} & 25.8 & 73.5 & 3.4 & \textcolor{red}{\textbf{3.0}} & \textcolor{red}{\textbf{31.5}} &19.5 & 21.3 & 67.4 & 69.4 & \textcolor{red}{\textbf{68.5}} & \textcolor{red}{\textbf{25.0}} & \textcolor{red}{\textbf{76.5}} &  \textcolor{red}{\textbf{41.6}} & 17.9 & 29.5 & 41.2 \\	
		\hline
		\hline		
		BL & 22.5 & 15.4 & 74.1 & 9.2 & 0.1 & 24.6 & 6.6 & 11.7 & 75.0 & 82.0 &56.5 & 18.7 & 34.0 & 19.7 & 17.1 & 18.5 & 30.4 \\				
		\hline			
		BL+AL & 74.4 & 30.5 & 75.8 & 13.2 & 0.2 & 19.7 & 4.4 &4.9 & 78.2 &\textcolor{red}{\textbf{82.7}} &44.4 &16.0 & 63.2 & \textcolor{blue}{33.3} & 13.5 &26.2 & 36.3  \\
		\hline	
		BL+AL+PL & 79.2 & \textcolor{blue}{38.7} & 76.5 & 10.7 & 0.3 & 22.4 & 5.6 & 11.4 & 79.5 & 81.3 & 58.1 & 20.7 & 70.4 & 31.6 & 24.8 & 32.3 & 40.2    \\
		\hline
		BL+AL+SRT & 79.9 & 38.2 &77.1 & 9.7 & 0.2& 21.1 &6.8 &7.6 &76.1 &81.6 &54.8 &\textcolor{blue}{21.3} &66.2 &30.8 &21.6 &30.6 & 39.0 \\
		\hline
		BL+PL+SRT & 61.6 & 28.7 & 71.6 & 20.8 &0.6 & 28.7 & \textcolor{red}{\textbf{31.1}} & \textcolor{blue}{24.9} & 80.0 & 81.5& 62.7 & 16.2 & 69.4 &12.3 & \textcolor{red}{\textbf{27.8}} & \textcolor{blue}{51.5} & 41.8 \\			
		\hline
		Ours-woSP & 67.2 & 29.4 & 73.5 & \textcolor{blue}{21.2} & \textcolor{blue}{0.7} & 28.4 & \textcolor{blue}{29.7} & 24.5 & 79.9 & 81.1 & 62.9 & 15.8 & 72.8 & 12.6 & \textcolor{blue}{26.5} & 51.2 & \textcolor{blue}{42.3} \\
		\hline
		Ours & 68.4 & 30.1 & 74.2 & \textcolor{red}{\textbf{21.5}} & 0.4 & \textcolor{blue}{29.2} & 29.3 &\textcolor{red}{\textbf{25.1}} & \textcolor{blue}{80.3} & 81.5 &\textcolor{blue}{63.1}& 16.4 & \textcolor{blue}{75.6} & 13.5 & 26.1 & \textcolor{red}{\textbf{51.9}} & \textcolor{red}{\textbf{42.9}}  \\			
		\hline 		
	\end{tabular}
	\vspace{-6.5pt}		
	\caption{Comparisons performance of learning transferable knowledge from SYNTHIA dataset to Cityscapes dataset. Models with best and runner-up performance are marked with red and blue colors, respectively.} 
	\label{table:exp_synthia_cityscapes}	
\end{table*}

\begin{table*}[t]
	\vspace{5pt}
	\scriptsize
	\centering
	\setlength{\tabcolsep}{1.24mm}
	\begin{tabular}{|c|c|c|c|c|c|c|c|c|c|c|c|c|c|c|c|c|c|c|c|c|}
		\hline
		Method & road & sidewalk & building & wall & fence & pole &light &sign & veg & terrain & sky  & person & rider & car & truck & bus & train & mbike & bike & mIoU \\
		\hline	
		\hline
		DF \cite{exp:DF} & 31.9 & 18.9 & 47.7 & 7.4 & 3.1 & 16.0& 10.4 & 1.0 &76.5 & 13.0 & 58.9 & 36.0 & 1.0 & 67.1 & 9.5 & 3.7 & 0.0 & 0.0& 0.0 & 21.1\\
		\hline
		Wild \cite{DBLP:journals/corr/HoffmanWYD16} & 70.4 & 32.4 & 62.1 & 14.9 & 5.4 & 10.9 & 14.2 & 2.7 & 79.2& 21.3 & 64.6 & 44.1 & 4.2 & 70.4 & 8.0 & 7.3& 0.0 & 3.5 & 0.0 & 27.1 \\
		\hline
		CL \cite{exp:CL} & 74.9 & 22.0 & 71.7 & 6.0 & 11.9 & 8.4 & 16.3 & 11.1 & 75.7 & 11.3 & 66.5 & 38.0 & 9.3 & 55.2 & 18.8 & 18.9 & 0.0 & 16.8 & 14.6 & 28.9\\
		\hline
		CyCADA \cite{domain:class-preserve-2} &79.1 &33.1 &77.9 &23.4 &17.3 &32.1 &33.3 &31.8 &\textcolor{blue}{81.5} &26.7 &69.0 &\textcolor{blue}{62.8} &14.7 &74.5 &20.9 &25.6 & 6.9&18.8 &20.4 & 39.5\\
		\hline
		LSD \cite{exp:LSD} &88.0 &30.5 &78.6 &25.2 &23.5& 16.7 &23.5 &11.6 &78.7 &27.2 &71.9 &51.3 &19.5 &80.4 &19.8 &18.3 &0.9 &20.8 &18.4 &37.1 \\
		\hline
		LtA \cite{exp:LtA} & 86.5 & 36.0 & \textcolor{red}{\textbf{79.9}} & 23.4 & 23.3 &23.9 &35.2 &14.8 & \textcolor{red}{\textbf{83.4}}& \textcolor{blue}{33.3}&  \textcolor{red}{\textbf{75.6}} & 58.5 & 27.6 & 73.7& \textcolor{blue}{32.5} & 35.4 & 3.9 & \textcolor{red}{\textbf{30.1}} & 28.1 & 42.4\\
		\hline
		CGAN \cite{exp:CGAN} & 89.2 & \textcolor{blue}{49.0} & 70.7&13.5 & 10.9& \textcolor{red}{\textbf{38.5}} & 29.4 & 33.7& 77.9 & \textcolor{red}{\textbf{37.6}} & 65.8 & \textcolor{red}{\textbf{75.1}} & \textcolor{red}{\textbf{32.4}}& 77.8 & \textcolor{red}{\textbf{39.2}}&45.2 & 0.0& 25.2 & \textcolor{red}{\textbf{35.4}} & 44.5 \\	
		\hline	
		\hline 		
		BL & 80.2 & 6.4 & 74.8 & 8.8 & 17.2 & 17.5 & 30.5 & 17.7 & 75.0 & 14.1 & 57.9 &56.2 & 27.3 &64.1 & 29.7 & 24.1 & 4.7 & \textcolor{blue}{27.6} & 33.4 & 35.1  \\  				
		\hline			
		BL+AL & 86.3 & 32.2 & \textcolor{blue}{79.8} &22.0 &22.2 &27.1 &33.5 &20.1 &80.3 & 21.5 &\textcolor{blue}{75.5} &59.0 &25.4 &73.1 & 28.0 &32.2 & 5.4 & 27.3& 31.5& 41.2 \\
		\hline
		BL+AL+PL & 91.7 & 48.3 & 76.8 & 25.1 & \textcolor{blue}{28.5} & 28.2 & 39.7 & 44.5 & 79.8 & 13.6 & 72.3 & 53.6 & 19.1 & 85.8 & 23.7 & 44.2 & 32.8 & 13.4 & 31.5 & 44.9  \\
		\hline
		BL+AL+SRT & 92.4 & \textcolor{red}{\textbf{49.8}} &73.6 &25.3 & 28.3 &24.5 &40.9 &45.0 &79.2 &14.2 & 70.4&50.1 & 18.6 & \textcolor{red}{\textbf{86.6}} & 22.3 & \textcolor{blue}{45.4} & 30.3 & 11.9 & 32.8 & 44.3 \\
		\hline	
		BL+PL+SRT & \textcolor{blue}{92.6} & 47.8 &77.4 &\textcolor{red}{\textbf{26.7}} &\textcolor{red}{\textbf{28.8}} &29.9 &\textcolor{red}{\textbf{42.4}} &\textcolor{red}{\textbf{46.3}} &80.7 &15.1 &71.1 &55.8 &24.3 &\textcolor{blue}{86.5} &21.5 &42.4 & \textcolor{red}{\textbf{43.3}} &12.1 &30.8 & 46.1 \\			
		\hline
		Ours-woSP & 92.4 & 47.3 & 78.5 & 25.4 & 27.8 & 34.8 & 42.0 & 44.6 & 79.8 & 15.3 & 67.1 & 60.5 & \textcolor{blue}{30.7} & 86.3 & 26.4 & 43.7 & \textcolor{blue}{36.1} & 14.8 & 33.2 & \textcolor{blue}{46.7} \\
		\hline	
		Ours &\textcolor{red}{\textbf{92.7}} &48.0 &78.8 & \textcolor{blue}{25.7} & 27.2 & \textcolor{blue}{36.0} & \textcolor{blue}{42.2} & \textcolor{blue}{45.3}& 80.6 &14.6 &66.0 & 62.1& 30.4 &86.2 & 28.0 &\textcolor{red}{\textbf{45.6}} &35.9 &16.8 &\textcolor{blue}{34.7} & \textcolor{red}{\textbf{47.2}} \\		
		\hline 		
	\end{tabular}		
	\hspace{5pt}
	\vspace{-7pt}	 
	\caption{Comparison performance of learning transferable representation from GTA dataset to Cityscapes dataset. Models with best and runner-up performance are marked with red and blue colors, respectively.}	
	\label{table:exp_gta_cityscapes}
	\vspace{-2pt}
\end{table*}

\vspace{-5pt}
\subsection{Experiments on Benchmark Datasets} 	
In this subsection, we conduct experiments on several benchmark datasets that has compatible annotations with each other to further justify the effectiveness of our model. For a fair comparison, we remove the classification head and adopt the same experimental data configuration with the completing methods \cite{DBLP:journals/corr/HoffmanWYD16, exp:CL, domain:class-preserve-2, exp:LSD, exp:LtA,exp:CGAN}. For the ablation studies shown in Table~\ref{table:exp_synthia_cityscapes} and Table~\ref{table:exp_gta_cityscapes}, BL, AL, PL, SRT and Ours-woSP indicate baseline, adversarial learning, pseudo labels, semantic lesions transfer components of our model and training without super-pixel priors, respectively.

\textbf{Transfer from SYNTHIA to Cityscapes:} In this experiment, our model is used to learn transferable knowledge from SYNTHIA \cite{data:synthia} to Cityscapes \cite{data:city}. For the training phase, SYNTHIA dataset with finely-annotated 9400 images is regarded as $X^s$. The Cityscapes without pixel labels has 2993 images is regarded as $X^t$. For the test, we use validation subset with 500 images of Cityscapes, which is disjoint with training subset. Notice that we consider 16 common classes for two datasets: road, sidewalk, building, wall, fence, pole, traffic light, traffic sign, vegetation, sky, person, rider, car, bus, motorbike and bike.  From the presented results in Table~\ref{table:exp_synthia_cityscapes}, we can conclude that: 1) Our model outperforms state-of-the-arts \cite{exp:LSD, exp:CGAN, exp:LtA} by $1.7\sim6.8\%$ for the remaining classes in terms of mIoU, which verifies the effectiveness of our model; 2) Ablation studies of both PL, SRT and SP also validates these components are designed reasonably; 3) Although the appearances of the hard-to-transfer classes (e.g., wall, pole, motorbike and bike) are extremely different between these two datasets, our model can also achieve comparable performance.

\textbf{Transfer from GTA to Cityscapes:} When conducting experiments to learn transferable representation from GTA \cite{data:GTA} to Cityscapes \cite{data:city}, in the training process, GTA with finely-annotated 24996 images and the training subset with 2993 images of Cityscapes without using pixel labels are treated as $X^s$ and $X^t$, respectively. The remaining validation subset with 500 images of Cityscapes is used for evaluation. As the results presented in Table~\ref{table:exp_gta_cityscapes}, we consider 19 shared classes: road, sidewalk, building, wall, fence, traffic light, traffic sign, vegetation, terrain, sky, person, rider, car, truck, bus, train, motorbike and bike. Notice that: 1) Other semantic transfer models can be easily partial towards easy-to-transfer classes (e.g., road, building, sky, vegetation and car), while our model can achieve better performance for both initially hard-to-transfer classes and easy-to-transfer classes. 2) The ablation studies of PL, SRT and SP verify that previously-learned experience and pseudo labels play a significant role when comparing with \cite{domain:class-preserve-2, exp:LSD, exp:LtA, exp:CGAN}.

\section{Conclusion}
\vspace{-3pt}	
In this paper, we explore a new semantic lesions representation transfer model for weakly-supervised endoscopic lesions segmentation. More specifically, a pseudo pixel label generator is presented to progressively mine more samples from target data into training set, which incorporates super-pixel priors and class balance to prevent dominance of well-to-transfer categories. We also align the  dynamically-searched feature centroids for each class of different datasets with previously-learned features. Experiments on our built dataset and several benchmark datasets show the effectiveness and superiority of our model.  

\vspace{-4pt}
\section*{Acknowledgment}
\vspace{-2pt}
Thanks for the medical endoscopic data provided by Prof. Yunsheng Yang from the Chinese PLA General Hospital.

{\small
\bibliographystyle{ieee_fullname}
\bibliography{LesionTransfer} 
}

\end{document}